\documentclass{article} 
\usepackage{baichuan}

\usepackage[utf8]{inputenc} 
\usepackage[T1]{fontenc}    
\usepackage{hyperref}       
\usepackage{url}            
\usepackage{booktabs}       
\usepackage{amsfonts}       
\usepackage{nicefrac}       
\usepackage{microtype}      
\usepackage{xcolor}         
\usepackage{multirow}
\usepackage{multicol}
\usepackage{amsmath}
\usepackage{graphicx}

\usepackage{subcaption}
\usepackage{utfsym}
\usepackage{svg}
\usepackage{array}
\usepackage{fancyhdr}
\usepackage{tablefootnote}
\usepackage{footnote}

\usepackage{hyperref}
\usepackage{orcidlink}
\usepackage{makecell}
\usepackage{color}
\usepackage{tcolorbox}
\usepackage{colortbl}
\usepackage{pifont}
\usepackage{enumitem}
\usepackage{dashrule}
\usepackage{bbm}
\newcommand{\ignore}[1]{}
\newcommand{\ie}{\emph{i.e.,} }

\newcommand{\eg}{\emph{e.g.,} }

\definecolor{kellygreen}{rgb}{0.3, 0.73, 0.09}
\definecolor{alizarin}{rgb}{0.82, 0.1, 0.26}

\title{Exploring the Design Space of Visual Context Representation in Video MLLMs}

\author{
	\begin{tabular}[t]{c}
		Yifan Du\textsuperscript{1,2}\thanks{Equal contribution.}, 
		Yuqi Huo\textsuperscript{2}*,
        Kun Zhou\textsuperscript{3}*,
        Zijia Zhao\textsuperscript{4},
        Haoyu Lu\textsuperscript{1},
        Han Huang\textsuperscript{4},\\
		Wayne Xin Zhao\textsuperscript{1}\thanks{Corresponding authors.},
		Bingning Wang\textsuperscript{2$\dagger$},
		Weipeng Chen\textsuperscript{2}, 
		and Ji-Rong Wen\textsuperscript{1,3}
		\\
		\textnormal{\textsuperscript{1}Gaoling School of Artificial Intelligence, Renmin University of China} \\
        \textnormal{\textsuperscript{2}Baichuan Inc.} \\
        \textnormal{\textsuperscript{3}School of Information, Renmin University of China} \\
        \textnormal{\textsuperscript{4}Institute of Automation, Chinese Academy of Sciences}\\
        \texttt{\{yifandu1999, batmanfly\}@gmail.com}, 
        \texttt{daniel@baichuan-inc.com}
	\end{tabular}
	\\
}

\colmfinalcopy

\begin{document}	
\maketitle

\begin{abstract}
Video Multimodal Large Language Models~(MLLMs) have shown remarkable capability of understanding the video semantics on various downstream tasks. Despite the advancements, there is still a lack of systematic research on visual context representation, which refers to the scheme to select frames from a video and further select the tokens from a frame. In this paper, we explore the design space for visual context representation, and aim to improve the performance of video MLLMs by finding more effective representation schemes. Firstly, we formulate the task of visual context representation as a constrained optimization problem, and model the language modeling loss as a function of the number of frames and the number of embeddings (or tokens) per frame, given the maximum visual context window size. Then, we explore the scaling effects in frame selection and token selection respectively, and fit the corresponding function curve by conducting extensive empirical experiments. We examine the effectiveness of typical selection strategies and present empirical findings to determine the two factors. Furthermore, we study the joint effect of frame selection and token selection, and derive the optimal formula for determining the two factors. We demonstrate that the derived optimal settings show alignment with the best-performed results of empirical experiments. Our code and model are available at: \url{https://github.com/RUCAIBox/Opt-Visor}.
    
\end{abstract}

\section{Introduction}
\label{sec:intro}

Recent advancements in video Multimodal Large Language Models (\emph{video MLLMs}) have shown the great potential in extending LLMs to process video data~\citep{lin2023video}. 
Typically, a video MLLM is developed based on a pre-trained LLM, and an image encoder will be attached to the LLM via a modality projector, which links the textual and visual semantic spaces. In this way, we can prompt the video MLLM with textual instruction and visual embeddings, to generate the natural language response for fulfilling the video-based task, \eg video question answering~\citep{xu2017video} and video captioning~\citep{caba2015activitynet}.
Despite the success, it is still challenging for existing video MLLMs to handle complex or long videos, due to the limited model capacities. 


To develop effective video MLLMs, previous research work mainly focuses on two aspects, either improving the model architecture~\citep{wang2024longllava} or enhancing the model training~\citep{zhang2024direct, liu2024kangaroo}. However, another important aspect has been missing in the related literature, \ie visual context representation. In this work, \emph{visual context} refers to the visual embeddings in the prompt of video MLLMs. Unlike text and images, it is not very straightforward to represent a video. In existing approaches, a widely used way is to sample a number of frames from a video (\emph{frame selection}) and then further sample or generate a number of embeddings for each selected frame (\emph{embedding selection}). However, it is unclear how each factor affects the performance of video MLLMs, and how both factors jointly contribute to the performance improvement within the limited context length of the underlying LLM.

Considering this issue, in this paper, we take the initiative to explore the design space for visual context representation, and derive more effective representation schemes to improve the performance of video MLLMs. 
Specifically, we firstly formulate the studied task as a constrained optimization problem: given the maximum visual context window size, we model the language modeling loss as a function of the number of frames and the number of embeddings (or tokens) per frame. 
Such a formulation is useful to help understand the competitive relationships between frame selection and embedding selection.  
Subsequently, we conduct extensive empirical experiments to explore the scaling effects in frame and embedding selection respectively, and fit the corresponding function to describe the performance trend. 
Our findings show that: (1) overall increasing the number of visual embeddings (either tokens or frames) would enhance the performance, while scaling the frames can lead to consistently improved performance; (2)  
 the compression-based method can effectively preserve more semantic information with fewer visual embeddings. 
Furthermore, we study the joint effect of the two factors and propose the method to find the optimal allocation given the limited context length, which is further supported by empirical experiments.

The major contributions of our work are as follows:

\begin{itemize}[left=0pt]

\item To the best of our knowledge, this is the first work to systematically study the design of visual context, which is an important yet under-explored problem for developing capable video MLLMs. We provide both theoretical formulations and empirical findings to approach this problem. We also release a model under the \textbf{Opt}imal \textbf{Vis}ual C\textbf{o}ntext \textbf{r}epresentation scheme, \textbf{Opt-Visor},  which can process videos up to 162 frames.


\item We study the scaling effects of model performance \emph{w.r.t.} the number of selected frames and the number of selected embeddings per frame respectively. We fit the corresponding function curve, and compare different strategies (\ie sampling- and compression-based methods) for both factors.    



\item We explore the trade-off relationships for frame and embedding selection, and suggest the optimal formula for determining the two factors. We demonstrate that the derived optimal settings show alignment with the best-performed results of empirical experiments.




\end{itemize}

\section{Preliminary}
\label{sec:method}

In this section, we introduce the background for building the base model in our work. 

\paragraph{Model Architecture.} Following existing works~\citep{liu2024visual, zhang2024llavanextvideo}, we adopt the LLaVA-like model architecture,  consists of a visual encoder, an LLM, and a projector that maps the visual embeddings to the semantic space of the LLM. Formally, given a video with $T$ frames $\{I_{t}\}_{t=1}^T$, each frame is encoded by the image encoder $f_{\phi}$ to obtain $M$ visual embeddings $\{\textbf{v}^t_i\}_{i=1}^M$,  
\ignore{\begin{equation}
    \{\textbf{v}^t_i\}_{i=1}^M = f_{\phi}(I_t), 
\end{equation}}
where $\textbf{v}^t_i \in \mathbb{R}^{d_v}$ denotes the $i$-th visual embedding in the $t$-th frame, and $d_v$ is the dimensionality of the visual embedding. Then the projector $f_{\psi}$ projects these visual embeddings into the semantic space of the LLM, producing $\textbf{h}_i^t \in \mathbb{R}^d$. 
\ignore{\begin{equation}
    \textbf{h}_i^t=f_{\psi}(\textbf{v}_i^t), 
\end{equation}}
These visual embeddings are concatenated with the embeddings of a textual prompt $\{\textbf{e}_j\}_{j=1}^N$, where $\textbf{e}_j\in \mathbb{R}^d$ is the embedding of the $j$-th token in the prompt. The concatenated sequence is fed as input to the LLM $f_{\theta}$ to generate the output:
\begin{equation}
    y_1\cdots y_K = f_{\theta}([\textbf{h}_1^1, ..., \textbf{h}_M^1,...,\textbf{h}_1^T,...,\textbf{h}_M^T, \textbf{e}_1, ..., \textbf{e}_N])
\end{equation}
During training, we optimize the parameters $\{\phi, \psi,\theta\}$ by minimizing the next-token prediction loss. 

\ignore{\subsection{Model Architecture}
Following existing works~\citep{liu2024visual, zhang2024llavanextvideo}, we adopt the LLaVA-like model architecture. Specifically, our model consists of a visual encoder, an LLM, and a projector that maps the visual embeddings to the semantic space of the LLM. Formally, given a video with $T$ frames $\{I_{t}\}_{t=1}^T$, each frame is encoded by the image encoder $f_{\phi}$ to obtain $M$ visual embeddings: 
\begin{equation}
    \{\textbf{v}^t_i\}_{i=1}^M = f_{\phi}(I_t), t=1,2,...,T
\end{equation}
where $\textbf{v}^t_i \in \mathbb{R}^{d_v}$ denotes the $i$-th visual embedding in the $t$-th frame, and $d_v$ is the dimensionality of the visual embedding. Then the projector $f_{\psi}$ projects these visual embeddings into the semantic space of the LLM, producing $\textbf{h}_i^t \in \mathbb{R}^d$:
\begin{equation}
    \textbf{h}_i^t=f_{\psi}(\textbf{v}_i^t), i=1,2,...,M; t=1,2,...,T
\end{equation} 
These visual embeddings are concatenated with the embeddings of a textual prompt $\{\textbf{e}_j\}_{j=1}^N$, where $\textbf{e}_j\in \mathbb{R}^d$ is the embedding of the $j$-th token in the prompt. The concatenated sequence is used as input to the LLM $f_{\theta}$, which generates the output sequence $\{{y_k}\}_{k=1}^K$:
\begin{equation}
    \{{y_k}\}_{k=1}^K = f_{\theta}([\textbf{h}_1^1, ..., \textbf{h}_M^1,...,\textbf{h}_1^T,...,\textbf{h}_M^T, \textbf{e}_1, ..., \textbf{e}_N])
\end{equation}
During training, we optimize the parameters $\{\phi, \psi,\theta\}$ by minimizing the next-token prediction loss.
}

\begin{figure}
    \centering
    \includegraphics[width=1\linewidth]{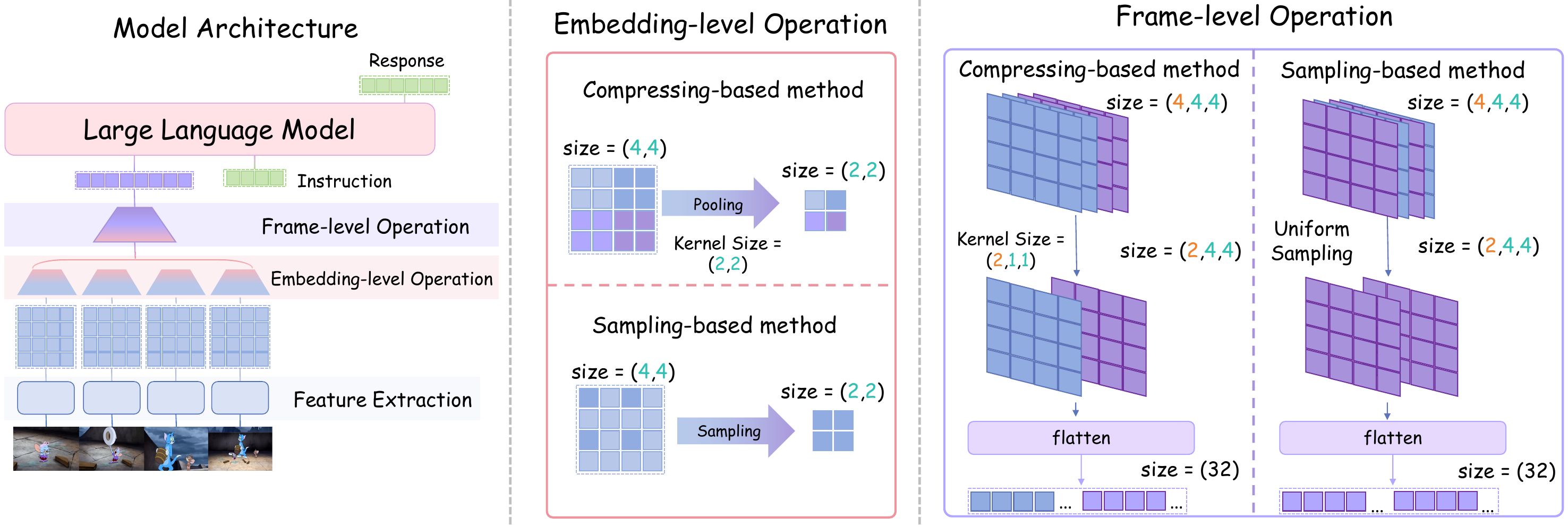}
    \caption{Overview of the LLaVA-like architecture for video-MLLM, and our used frame-level and embedding-level operations for adjusting the visual context window size.}
    \label{fig:model}
\end{figure}

\paragraph{Training Data.}
Based on existing instruction datasets, we mix several widely used image instruction and video instruction sets to construct a new instruction dataset. For the image instruction set, we  adopt Cauldron~\citep{laurenccon2024matters}, which is a large image instruction set based on 50 vision-language datasets. For the video instruction set, we collect the instructions from VideoChatGPT-100K~\citep{Maaz2023VideoChatGPT}, ShareGPT4Video~\citep{chen2024sharegpt4video}, ShareGPTVideo~\citep{zhang2024direct}, VIM~\citep{du2024towards}, as well as some instruction data from VideoChat2~\citep{li2024mvbench}. The statistics of each instruction set are listed in Table~\ref{tab:dataset}. 

\ignore{
\subsection{Training Data}
Based on existing instruction datasets, we mix several widely used image instruction and video instruction sets to construct a new instruction dataset. For the image instruction set, we  adopt Cauldron~\citep{laurenccon2024matters}, which is a large image instruction set based on 50 vision-language datasets. For the video instruction set, we collect the instructions from VideoChatGPT-100K~\citep{Maaz2023VideoChatGPT}, ShareGPT4Video~\citep{chen2024sharegpt4video}, ShareGPTVideo~\citep{zhang2024direct}, VIM~\citep{du2024towards}, as well as some instruction data from VideoChat2~\citep{li2024mvbench}. The statistics of each instruction set are listed in Table~\ref{tab:dataset}.
}

\paragraph{Implementation Details.}
We adopt SigLIP~\citep{zhai2023sigmoid} as the image encoder, Qwen2-7B~\citep{yang2024qwen2} as the base LLM, and a {two-layer} MLP as the projector. We train all the models with the training data listed in Table~\ref{tab:dataset} for 1 epoch. We have tried to include a pre-training stage before the visual instruction tuning, using the 558K pre-training data and only updated the parameters in the MLP following LLaVA~\citep{liu2024visual}, but found no obvious difference. 
All the experiments are conducted on 32 Nvidia H800, with the detailed hyperparameters listed in Table~\ref{tab:hyperparameter}.

\paragraph{Evaluation Setup}
To quantitatively assess the scaling effect of visual context in video MLLMs, we consider the following two metrics for evaluation: 

$\bullet$ \emph{Language Modeling Loss.} 
It is a continuous measure that reflects model performance in predicting the next token, which can be used to estimate the parameters of the scaling law function. 
To align with the setting in Chinchilla~\citep{hoffmann2022training}, each model will be trained for 1 epoch, {to ensure that training samples have not been seen when we calculate the loss for evaluation. } 


$\bullet$ \emph{Zero-shot Accuracy.} The zero-shot accuracy can reflect the performance of the model in the real-world application. 
We select several long video understanding benchmarks for evaluation, including  Event-Bench~\citep{du2024towards} (only with the challenging episodic reasoning task), NBench~\citep{zhao2024needle}, MLVU~\citep{zhou2024mlvu}, and VideoMME~\citep{fu2024video}. All the questions in these benchmarks are multiple-choice, and we use accuracy as the evaluation metric. 

\section{Scaling law of visual context}
\label{sec:scaling}

\subsection{Problem Formulation}

As introduced in Section~\ref{sec:method}, existing video LLMs typically follow the vision-language model architecture~\citep{liu2024visual, zhang2024llavanextvideo}, which represents a video into multiple representative frames. 
Further, each frame will be encoded into a number of visual  tokens or embeddings. The aggregation of  the visual embeddings from all selected frames is referred to as \emph{visual context} in this work. 
To set the visual context, it is essential to determine two aspects when the base architecture is fixed: 
\ignore{\begin{itemize}
\item \emph{frame selection}: how to select the frames from a video;
\item \emph{embedding selection}: how to select the visual tokens from an input frame. 
\end{itemize}} 
(1) how to select the frames from a video  (\emph{frame selection}), and (2) how to select the visual embeddings from an input frame (\emph{embedding selection}). 
Since the base architecture is developed on an existing LLM, the length of visual context is naturally limited by its  context length, \ie the maximum length of input tokens. The two aspects would be competitive in input length allocation: the more the selected frames, the fewer the visual embeddings per selected frame, and vice versa. 

In this work, we study the optimal allocation relationship of the visual context for a given video. Formally, we model the language modeling loss $\mathcal{L}(T,M)$ as a function of the number of frames $T$ and the number of embeddings (or tokens) per frame $M$. Given  the maximum visual context window size  $L$, the number of frames $T$ and visual embeddings per frame $M$ should satisfy the constraint: $T\times M<L$, we aim to find the optimal solution in minimizing $\mathcal{L}(T,M)$ under this constraint:
\begin{equation}
T_{\text{opt}}(L), M_{\text{opt}}(L) = \underset{T,M \text{ s.t. } T \times M < L}{\operatorname{arg min}} \mathcal{L}(T,M),
\label{equ:scaling_loss}
\end{equation}
where $T_{\text{opt}}(L)$ and $M_{\text{opt}}(L)$ represent the optimal allocation strategy for the frame and visual embedding, respectively, with the input limit $L$.
To approach it, in the following, we will explore the scaling effect of the number of frames and embeddings in Section~\ref{sec:vary_tokens} and Section~\ref{sec:vary_frames} respectively.





\subsection{Scaling Effect of the Visual Embeddings}
\label{sec:vary_tokens}
We first analyze the scaling effect of visual embeddings in a frame for a fixed number of frames. Specifically, we utilize two methods to select (or generate) the visual embeddings in a frame: the sampling- and compression-based method.

\subsubsection{Sampling-based Method}
\label{sec:sampling-token}

\paragraph{Experimental Setup.}
In this part, each image is first converted to $27\times 27$ embeddings by the image encoder, then we vary the number of sampled visual embeddings. Specifically, we uniformly sample $\{1^2, 2^2, 3^2, 4^2, 5^2, 6^2, 7^2, 9^2, 14^2\}$ embeddings from the $27\times 27$ embeddings, as illustrated in Figure~\ref{fig:model}. Other sampling methods like {block sampling~\citep{li2023inverse}} will be explored in future work. We set $T=32$ as the constant, and uniformly sample frames from each video and keep all other factors the same to train 9 video MLLMs with this setup of visual embeddings.

\paragraph{Fitting Function.}
We propose the following function to fit the scaling law of visual embeddings:
\begin{equation}\label{equ:scaling_token}
    \mathcal{L}(M) = L_M + \left(\frac{M_0}{M}\right)^{\alpha_M}
\end{equation}

\begin{figure}[h]
  \centering
  \begin{subfigure}[b]{0.45\linewidth}
    \centering
    \includegraphics[width=\linewidth]{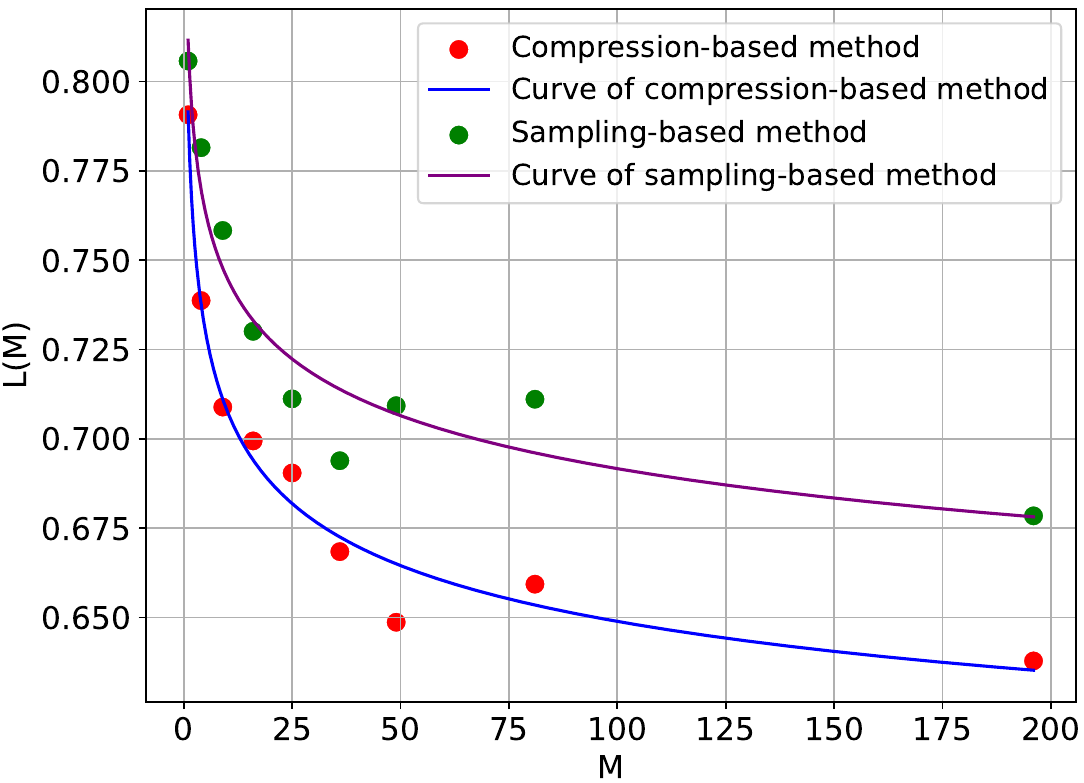}
    \caption{The scaling curve of visual embeddings.}
    \label{fig:fit_token_curve_both}
  \end{subfigure}
  \hspace{0.05\linewidth} 
  \begin{subfigure}[b]{0.45\linewidth}
    \centering
    \includegraphics[width=\linewidth]{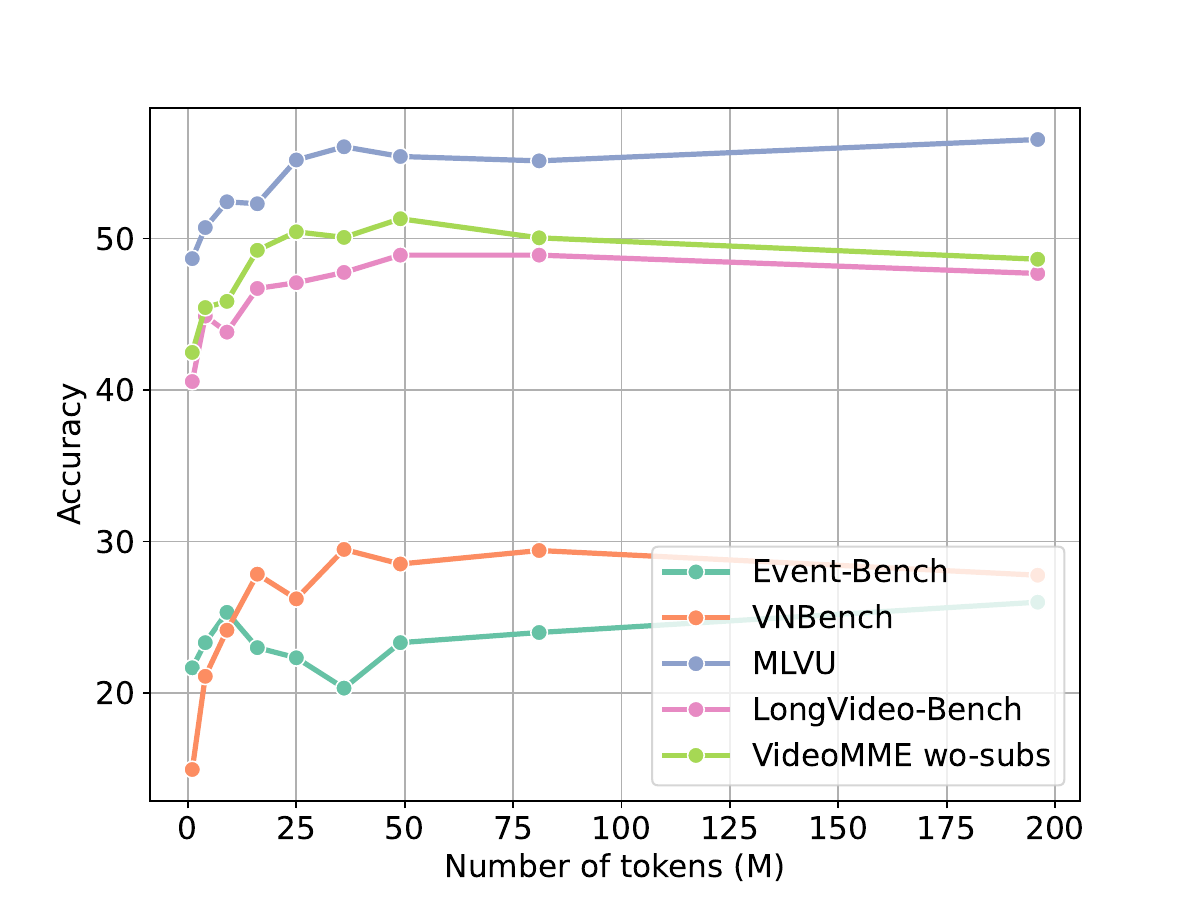}
    \caption{The relationship between the number of embeddings per frame and the benchmark accuracy. }
    \label{fig:benchmark_acc_token}
  \end{subfigure}
  \caption{The scaling law of visual embeddings, reflected by the language modeling loss and the zero-shot accuracy on video understanding benchmarks.}
  \label{fig:comparison_token}
\end{figure}

We fit the language modeling loss with respect to the number of visual embeddings $M$ using the scipy curvefit function, obtaining  $L_M = 0.48, M_0 = 1.16\times 10^{-5}, \alpha_M = 0.1$, with $R^2 = 0.927$ indicating a good fit. The fit curve in Figure~\ref{fig:fit_token_curve_both} shows that $\mathcal{L}(M)$ decreases with increasing $M$, following a power-law like trend. {We calculate the mean squared error between the actual loss and predicted loss, obtaining a value of 0.0001, which indicates a very low fitting error.}

\paragraph{Benchmark Performance Analysis}
Table~\ref{tab:sampling_token_result} shows the results of scaling visual embeddings on the evaluation benchmarks. Overall, the model performance improves as the number of visual embeddings increases, especially when it varies from 1 to 4. The improvement becomes more marginal with increasingly more visual embeddings. 
However, when it exceeds some threshold, the performance starts to decrease. For example, using 196 tokens is worse than using 49 tokens. An interesting finding is that the language modeling loss with 196 embeddings is significantly smaller than that of the model trained with 49 embeddings, as shown in Figure~\ref{fig:fit_token_curve_both}, which indicates that model loss might not directly reflect the performance on downstream tasks.  


\begin{table}[t]
\caption{Results of sampling-based method under different number of visual embeddings per frame.}
    \small
    \centering
    \begin{tabular}{c>{\columncolor{gray!20}}c|ccccc>{\columncolor{gray!20}}c}
    \toprule
 \# Frames&\makecell[c]{\# Embed./\\Frames}&\makecell[c]{Event-\\Bench}&  \makecell[c]{VNBench}& \makecell[c]{MLVU}&  \makecell[c]{LongVideo-\\Bench} &\makecell[c]{VideoMME\\wo/w-subs}   &Avg.\\
    \midrule
         32& \cellcolor{gray!20}1& 21.67& 14.96& 48.67& 40.56&42.48/52.04&\cellcolor{gray!20}36.73\\
         32& \cellcolor{gray!20}4& 23.33& 21.11& 50.72& 44.88&45.44/54.81&\cellcolor{gray!20}40.05\\
         32&\cellcolor{gray!20}9&\underline{25.33}&  24.15&  52.42&  43.82&45.85/54.67&\cellcolor{gray!20}41.04\\
         32&\cellcolor{gray!20}16&23.00& 27.85& 52.29&  46.70&49.22/57.11&\cellcolor{gray!20}42.20\\
         32&\cellcolor{gray!20}25&22.33& 26.22& 55.18&  47.08&\underline{50.44}/\underline{58.26}&\cellcolor{gray!20}43.25\\
         32&\cellcolor{gray!20}36&20.33&  \textbf{29.48}&  \underline{56.05}&  \underline{47.76}&50.07/58.04&\cellcolor{gray!20}43.62\\
         32&\cellcolor{gray!20}49&23.33&  28.52&  55.41&  \textbf{48.90}&\textbf{51.30}/\textbf{59.74}&\cellcolor{gray!20}\textbf{44.53}\\
         32&\cellcolor{gray!20}81&24.00&  \underline{29.41}&  55.12&  \textbf{48.90}&50.04/57.30&\cellcolor{gray!20}\underline{44.13}\\
         32&\cellcolor{gray!20}196&\textbf{26.00}&  27.78&  \textbf{56.53}&  47.69&48.63/55.59&\cellcolor{gray!20}43.70\\
    \bottomrule
    \end{tabular}
    \label{tab:sampling_token_result}
\end{table}



\subsubsection{Compression-based Method}
\label{sec:compress-token}

\paragraph{Experimental Setup.}
We utilize the MeanPooling~\citep{yao2024deco} strategy for compressing the visual embeddings, which has been widely used in visual information processing. Another advantage is that it does not introduce extra parameters, avoiding the influence of new factors in the experiments.
We apply MeanPooling with different kernel sizes on the feature map produced by the image encoder and obtain the condensed representation of the image. Specifically, each image is encoded into $27\times 27$ visual embeddings, on which we apply $p\times p$ mean pooling with stride $p$ ($p=\{2,3,4,5,6,7,9,14,27\}$), obtaining $\{1^2, 2^2, 3^2, 4^2, 5^2, 6^2, 7^2, 9^2, 14^2\}$ condensed embeddings per image. All the other factors are kept the same for fair comparison.

\begin{table}[t]
\caption{Model performance under different numbers of visual embeddings per frame, with the compression-based method.}
    \small
    \centering
    \begin{tabular}{c>{\columncolor{gray!20}}c|ccccc>{\columncolor{gray!20}}c}
    \toprule
 \# Frames&\makecell[c]{\# Embed./\\Frames}&\makecell[c]{Event-\\Bench}&  \makecell[c]{VNBench}& \makecell[c]{MLVU}&  \makecell[c]{LongVideo-\\Bench} &\makecell[c]{VideoMME\\wo/w-subs}   &Avg.\\
    \midrule
         32& 1& 18.67& 18.44& 49.45& 39.88&41.33/49.15 &36.15\\
         32& 4& 24.33& 27.04& 52.77& 43.44&46.11/55.93 &41.60\\
         32&9&22.33&  27.41&  53.83&  45.11&47.78/55.41 &41.98\\
         32&16&20.33& 28.96& 55.04&  46.32&49.85/58.15 &43.11\\
         32&25&23.00& 28.67& 54.21&  46.02&49.85/58.11 &43.31\\
         32&36&\underline{27.33}&  \underline{30.00}&  53.73&  48.45&50.33/58.74 &\underline{44.76}\\
         32&49&21.33&  29.93&  54.84&  47.16&49.96/58.37 &43.60\\
         32&81&23.33&  27.33&  \textbf{57.59}&  \underline{48.60}&\underline{52.00}/\underline{59.37} &44.70\\
         32&196&\textbf{29.00}&  \textbf{31.56}&  \underline{56.81}&  \textbf{52.24}&\textbf{53.56/59.48} &\textbf{47.11}\\
    \bottomrule
    \end{tabular}
    \label{tab:compressing_token_result}
\end{table}

\paragraph{Fitting Function.}
We use Equation~\ref{equ:scaling_token} to fit the scaling law and obtain $L_M=0.57, M_0=0.01, \alpha_M=0.39$, with $R^2=0.987$. We also calculate the mean square error of the predicted loss and the actual loss, a value of $5.32\times 10^{-5}$ indicates a good fit. Compared to the parameters of sampling-based method in Section~\ref{sec:sampling-token}, where $\alpha_M=0.1$, the $\alpha_M$ of the compression-based method is significantly larger, implying that increasing the number of embeddings using the compression-based method will result in faster loss decrease, as shown in Figure~\ref{fig:fit_token_curve_both}. Additionally, the compression-based method consistently yields a lower loss than the sampling-based method for the same number of visual embeddings. Unlike sampling-based methods, compression-based method does directly discard  visual embeddings but instead aggregates information from them, which is useful to accelerate the optimization process. 


\paragraph{Benchmark Performance Analysis}
For the benchmark evaluation in Table~\ref{tab:compressing_token_result}, the overall accuracy consistently increases as the number of visual embeddings increases. This finding is significantly different from that in Table~\ref{tab:sampling_token_result}. 
This result highlights the advantage of the compression-based method, which can  preserve more information than the sampling-based method. The conclusion drawn from the benchmark evaluation aligns with that concluded from the language modeling loss.

\begin{center}
\begin{tcolorbox}[colback=blue!5!white,colframe=blue!55!black,width=1\textwidth,title={Take-away Findings}]
{\begin{itemize}[left=0pt]
    \item Increasing the number of visual embeddings can significantly enhance the performance, with the sampling-based method achieves the peak at 49 tokens while the compression-based method does not saturate even with 196 tokens.
    \item When the visual context window size is limited, the compression-based method can effectively preserve more visual information with fewer visual embeddings. 
\end{itemize}
}
\end{tcolorbox}
\end{center}

\subsection{Scaling Effect of the Selected Frames}
Next, we continue to explore the scaling effect of the selected frames by varying its number $T$ while fixing the number of embeddings per frame $M$.
We also consider utilizing sampling-based and compression-based methods. 
\label{sec:vary_frames}

\subsubsection{Sampling-based Method}

\begin{table}[t]
\caption{Experimental results under different number of frames, with the sampling-based method.}
    \small
    \centering
    \begin{tabular}{>{\columncolor{gray!20}}cc|ccccc>{\columncolor{gray!20}}c}
    \toprule
\makecell[l]{\# Frames} & \makecell[c]{\# Embed./\\Frames}&\makecell[c]{Event-\\Bench}&  \makecell[c]{VNBench}& \makecell[c]{MLVU}&  \makecell[c]{LongVideo\\Bench} &\makecell[c]{VideoMME\\wo/w-subs}  &Avg.\\
    \midrule
        8 &49&21.67&  15.70&  46.30&  44.73&44.85/52.74 &37.67\\
        16 &49&22.67& 23.33& 52.53&  46.78&49.74/57.59 &42.11\\
        32 &49&21.33& 29.93& 54.84&  47.16&49.96/58.37 &43.60\\
        48&49&22.67& 34.15& 56.22&  \underline{48.75}&52.81/59.11 &45.62\\
        64 &49&25.33&  32.59&  57.23&  
        47.08&52.59/58.93 &45.63\\
        96& 49& \textbf{26.67}& \underline{37.26}& \underline{60.97}& 48.60&\underline{53.26}/\underline{60.85} &\underline{47.94}\\
        128 &49&\underline{25.67}&  \textbf{39.70}&  \textbf{61.44}&  \textbf{51.40}&\textbf{56.11}/\textbf{61.63} &\textbf{49.33}\\
    \bottomrule
    \end{tabular}
    \label{tab:sampling_frame_result}
\end{table}

\paragraph{Experimental Setup.} 
In existing works, it has become a widely used practice to sample frames uniformly from the original video to accommodate for the context length of LLM. Based on this method, we sample different numbers of frames from the video to explore the scaling effect, by varying $T$ in $\{1,8,16,32,48,64,96,128\}$.
In Section~\ref{sec:trade-off}, we further increase $T$ to 162 to explore the limit of scaling frames. The maximum context length allowed by the computation memory is 8K, corresponding to 128 frames and 49 visual embeddings per frame.\ignore{Since the limited computation memory does not allow us to keep all the visual embeddings per frame and scale the number of frames from 1 to 128, we first reduce the number of embeddings per frame using the method in \ref{sec:vary_tokens} and keep it fixed, then scale the number of frames to conduct the experiment. \textcolor{blue}{According to the result in Table~\ref{tab:compressing_token_result}, using the compression-based method and setting $p=4$ reduces the number of visual embeddings per frame to 93\% of the original one, and maintains the performance on most benchmarks}.} As a result, we adopt $4\times 4$ MeanPooling to keep 49 embeddings per frame and train 8 video MLLMs by varying the number of frames from 1 to 128.

\paragraph{Fitting Function.}
Similarly, we use  the following function to fit the scaling law of frames:

\begin{equation}\label{equ:scaling_frame}
    \mathcal{L}(T) = L_T + \left(\frac{T_0}{T}\right)^{a_T}
\end{equation}

We fit the losses with the number of frames $T$ and obtain  $L_T = 0.14, T_0 = 5.37\times 10^{-7}, \alpha_T = 0.04$, with $R^2 = 0.892$. The fitted curve in Figure~\ref{fig:fit_token_curve_both} shows that $\mathcal{L}(T)$ decreases with increasing $T$, following a power-law like trend. {We calculate the mean squared error between the actual loss and predicted loss, obtaining a value of 0.0001, which indicates a very low fitting error.}

\begin{figure}[h]
  \centering
  \begin{subfigure}[b]{0.45\linewidth}
    \centering
    \includegraphics[width=\linewidth]{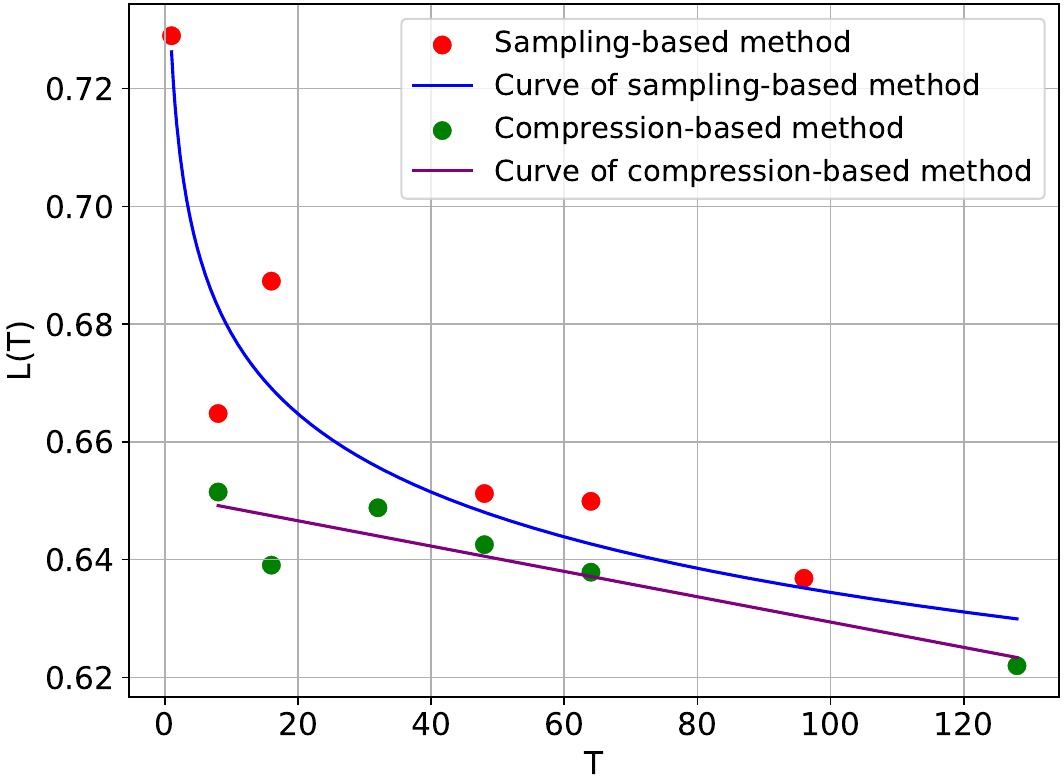}
    \caption{The scaling curve of frames.}
    \label{fig:fit_frame_curve_both}
  \end{subfigure}
  \hspace{0.05\linewidth} 
  \begin{subfigure}[b]{0.45\linewidth}
    \centering
    \includegraphics[width=\linewidth]{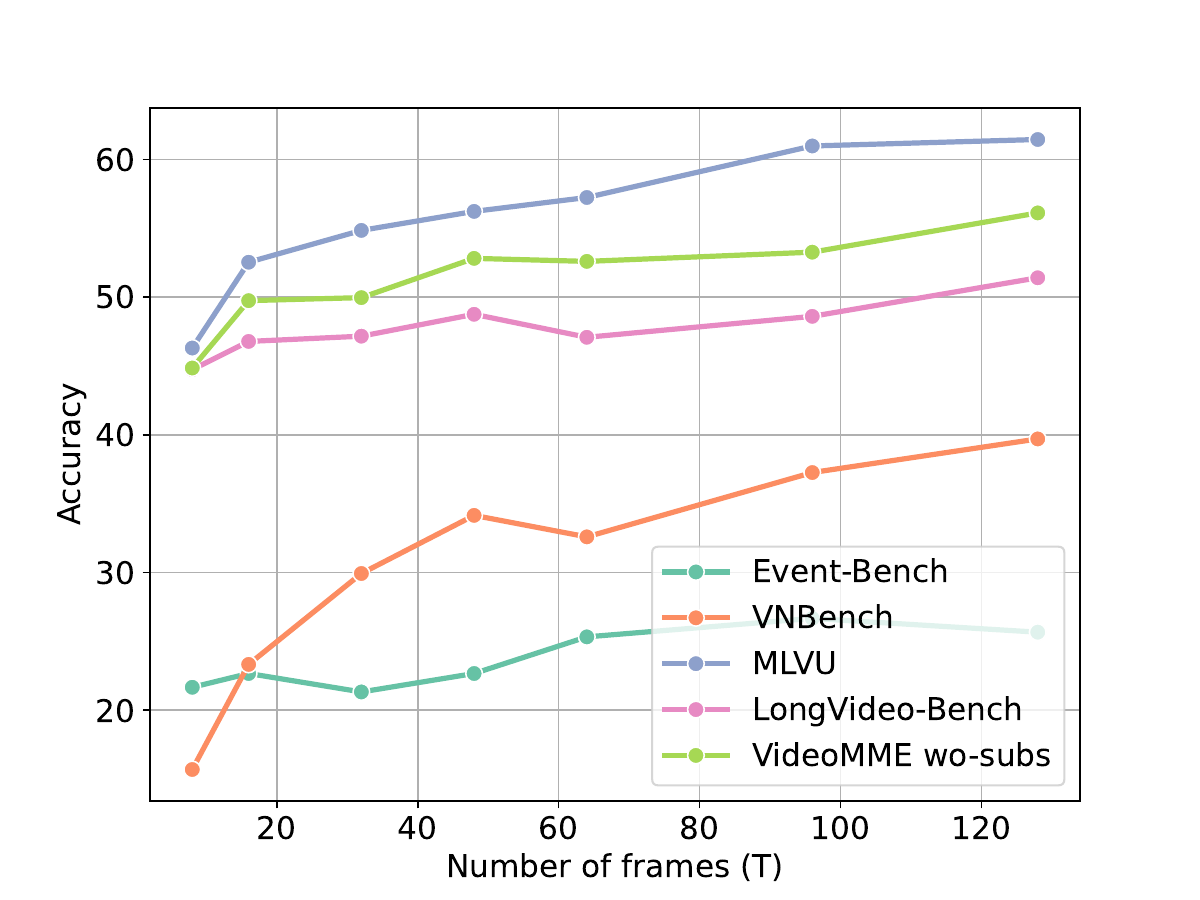}
    \caption{The relationship between the number of frames and the benchmark accuracy.}
    \label{fig:benchmark_acc_frame}
  \end{subfigure}
  \caption{The scaling law of frames, reflected by the language modeling loss and the zero-shot accuracy on video understanding benchmarks.}
  \label{fig:comparison_frame}
\end{figure}

\paragraph{Benchmark Performance Analysis}
The results in Table~\ref{tab:sampling_frame_result} and Figure~\ref{fig:benchmark_acc_frame} show that as the number of frames increases, the model consistently improves on all benchmarks, with no clear saturation point, even at 128 frames, which exceeds the maxinum frame count of most video MLLMs. Among all the benchmarks, VNBench shows that the most pronounced improvements (from 15.70 to 39.70), suggesting that the Needle-In-the-Haystack-Search (NIAH) task benefits most from extended temporal context. However, the Event-Bench shows no significant improvement beyond 64 frames, and a detailed inspection reveals that all questions in Event-Bench focus on episodic reasoning, a task that cannot be effectively learned by video MLLMs simply by increasing the number of frames~\citep{li2023llamavid}. Overall, compared to scaling the visual embeddings per frame (Figure~\ref{fig:benchmark_acc_token}), increasing the frames is more beneficial for improving the model performance.

\paragraph{Performance Compensating for Compressing Visual Embeddings.} {Another interesting finding is that the performance degradation caused by compressing visual embeddings can be compensated by increasing the number of frames. Specifically, Table~\ref{tab:compressing_token_result} shows that reducing the number of visual embeddings per frame from 196 to 49 leads to a performance drop across all benchmarks. However, if we simultaneously increase the number of frames to 128, the accuracy returns, even surpassing the model with 196 embeddings (comparing the last rows of Table~\ref{tab:compressing_token_result} and Table~\ref{tab:sampling_frame_result}, both setups use a total of 6272 visual embeddings, but one utilizes 32 frames with 196 embeddings per frame, while the other employs 128 frames with 49 embeddings per frame). These results suggest that when constrained by the visual context length, we can increase the number of frames while decreasing the embeddings per frame to achieve better performance, as will be further demonstrated in Section~\ref{sec:trade-off}.}

\begin{table}[t]
\caption{Model performance with the compression-based method under different numbers of frames. For the model trained with 128 frames, since $T_{\text{max}} = T = 128$ in this setting, the temporal pooling kernel size is $l = \lceil \frac{128}{128} \rceil = 1$, resulting in the same outcome as the sampling-based method.}
    \small
    \centering
    \begin{tabular}{>{\columncolor{gray!20}}cc|ccccc>{\columncolor{gray!20}}c}
    \toprule
\makecell[l]{\# Frames} & \makecell[c]{\# Embed./\\Frames}&\makecell[c]{Event-\\Bench}&  \makecell[c]{VNBench}& \makecell[c]{MLVU}&  \makecell[c]{LongVideo\\Bench} &\makecell[c]{VideoMME\\wo/w-subs}  &Avg.\\
    \midrule
        8 &49&25.67&  20.89&  53.04&  46.32& 50.48/57.41&42.30\\
        16 &49&\underline{27.00}& 27.04& 55.77&  48.07& 50.44/58.30&44.44\\
        32 &49&25.33& 29.78& 59.37&  \underline{48.52}& \underline{53.81}/\underline{60.93}&46.29\\
        48&49&24.33& \underline{37.41}& 59.31&  47.61& 52.07/59.81&46.76\\
        64 &49&\textbf{29.00}&  36.30&  \underline{61.03}&  
        47.61& 53.70/60.56&\underline{48.03}\\
        128$^*$ &49&25.67&  \textbf{39.70}&  \textbf{61.44}&  \textbf{51.40}& \textbf{56.11}/\textbf{61.63} &\textbf{49.33}\\
    \bottomrule
    \end{tabular}
    \label{tab:compressing_frame_result}
\end{table}

\subsubsection{Compression-based Method}

\paragraph{Experimental Setup.}
Compressing frames along the temporal dimension has been widely discussed in the field of video representation learning but remains underexplored in video MLLMs~\citep{cheng2024videollama}. Similar to the compression strategy used in Section~\ref{sec:vary_tokens}, we utilize MeanPooling here to reduce the number of frames input to the LLM. Specifically, we uniformly sample $T_{\text{max}}$ frames\footnote{If the original video duration $T^\prime \leq T_{\text{max}}$, we uniformly sample $T^\prime$ frames from it; otherwise, we uniformly sample $T_{\text{max}}$ frames, which is a common practice for video MLLMs.} from the video and encode them with the image encoder. Then, we apply MeanPooling along the temporal dimension to compress the video into $T$ frames, where the temporal pooling kernel size $l$ is determined by $T_{\text{max}}$ and $T$: $l=\lceil \frac{T_{\text{max}}}{T}\rceil$. Due to the limitation of computational memory, we set $T_{\text{max}}=128$ and $T=\{8,16,32,48,64,128\}$ to explore the scaling law, which is significantly larger than existing state-of-the-art video MLLMs that mostly use 32 or 64 frames as input~\citep{li2024llava,cheng2024videollama}. To ensure a fair comparison with the sampling-based method, we also reduce the number of visual embeddings per frame to 49. In practice, we utilize three-dimensional MeanPooling, instead of first performing spatial MeanPooling followed by temporal MeanPooling, to avoid over-smoothing of the feature maps.

\paragraph{Fitting Function.}
Different from the previous experiments, a power-law like function can't  fit the data points in this part. 
Instead,  we find that a simple linear function can well describe the function relationship, which is defined as follows: 


\begin{equation}
    \mathcal{L}(T) = a\times T + b 
\end{equation}

We fit the losses with the number of frames $T$ and obtain $a = -0.0002, b = 0.651$, with $R^2 = 0.807$. The fit curve is shown in Figure~\ref{fig:fit_frame_curve_both}. We calculate the mean squared error between the actual loss and predicted loss is $1.753\times 10^{-5}$, which indicates a very low fitting error.

\paragraph{Benchmark Performance Analysis}
Comparing the curve of the sampling-based method and the compression-based method, the latter always results in lower loss than the former. {This phenomenon reveals the temporal redundancy in video data, showing that temporal information can still be effectively preserved even when compressed into fewer frames.} The evaluation results on the benchmarks are shown in Table~\ref{tab:compressing_frame_result}. Overall, increasing the number of frames always leads to improved model performance. Additionally, compared with the sampling-based method, the compression-based method consistently achieves higher accuracy with the same number of frames. This aligns with the phenomenon that the compression-based method generally gives lower training loss as depicted in Figure~\ref{fig:fit_frame_curve_both}.

\begin{center}
\begin{tcolorbox}[colback=blue!5!white,colframe=blue!55!black,width=1\textwidth,title={Take-away Findings}]
\begin{itemize}[left=0pt]
    \item Increasing the number of frames consistently improves the performance, even compensating for the performance degradation caused by compressing visual embeddings per frame.
    \item When the visual context window size is limited, the compression-based method can preserve more temporal information than sampling-based method with fewer frames.
\end{itemize}
\end{tcolorbox}
\end{center}

\section{Trade-off between Visual Embeddings and Frames}
\label{sec:trade-off}

Section~\ref{sec:vary_tokens} and Section~\ref{sec:vary_frames} have discussed the scaling effect of visual embeddings and frames separately. 
In this section, we explore the joint effect of the two factors, and study the problem: \emph{how to jointly determine the numbers of visual embeddings and frames under the constraint of maximum input length of LLM or deployment resource?} Based on the theoretical and empirical analysis, we present a video MLLM under the \textbf{Opt}imal \textbf{Vis}ual C\textbf{o}ntext \textbf{r}epresentation scheme, \textbf{Opt-Visor}, which can process videos up to 162 frames.

\begin{figure}[t]
    \centering
    \includegraphics[width=0.7\linewidth]{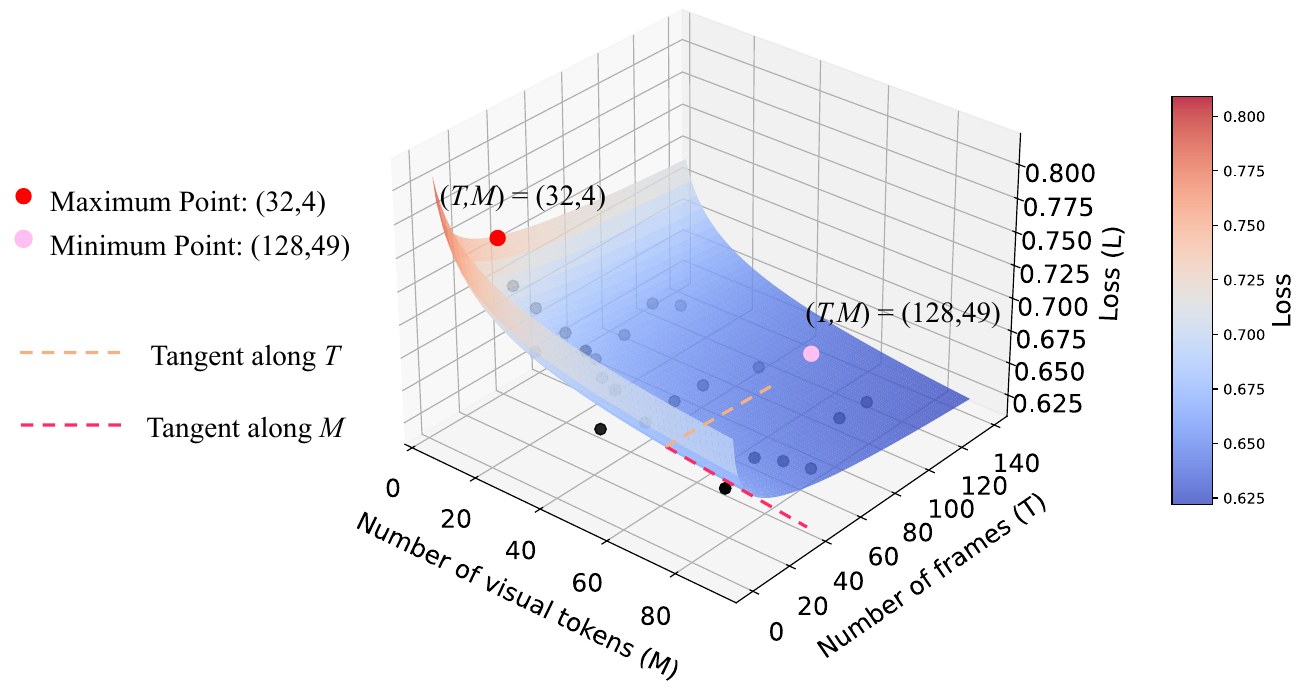}
    \caption{The scaling law of the visual embeddings and frames. We also display the maximum point and the minimum point in our experiments.}
    \label{fig:fit_both_curve_3d}
\end{figure}

\paragraph{Fitting Function of the Two Factors.} 
Following \citet{hoffmann2022training},  we fit the losses by considering the numbers of embeddings $M$ and frames $T$ as follows: 
\begin{equation}\label{equ:scaling_both}
    \mathcal{L}(M,T)=C_M\times M^{-\alpha}+C_T\times T^{-\beta}+L_0
\end{equation}
Specifically, we set the number of visual embeddings as $\{25,81\}$, and set the number of frames as $\{48,64,80,96\}$, train $2\times 4=8$ models in total. To extend the data points, we also include the 17 models trained in Section~\ref{sec:scaling}, and finally obtain 25 models in total. 
We obtain $C_M=0.25, \alpha=0.26, C_T=0.13, \beta=0.21, L_0=0.50$, with $R^2=0.884$. The fit curve along the axes of $T$ and $M$ is shown in Figure~\ref{fig:fit_both_curve_3d}. 
With the decreasing of both $T$ and $M$, the loss $\mathcal{L}(M,T)$ will consistently increase, reaching the highest loss at the data point $T=32, M=4$ in our experiment.
In contrast, as the $M$ and $T$ increase, the loss gradually decreases, and $T=128,m=49$ reaches the lowest loss. 
The computed gradient via the fitting function can help determine whether to increase $M$ or $T$ to achieve a lower loss. For example, the derivatives at $T=32, M=4$ are $\frac{\partial\mathcal{L}}{\partial M}=-0.01, \frac{\partial\mathcal{L}}{\partial T}=-0.004$, indicating that $\mathcal{L}$ descends faster along the $M$ direction. Therefore, increasing the number of embeddings is more promising to obtain lower loss, which aligns with our experiments.


    



\paragraph{Finding Optimal Setting.}
In practice, we are interested in the question that ``given the visual context window $L$, what is the best choice of $M$ and $T$ that achieves the lowest loss $\mathcal{L}(M,T)$''? To answer this question, we utilize the Lagrange multiplier method to obtain the minimum point of Equation~\ref{equ:scaling_both} under the constraint $M\times T <L$:

\begin{equation}\label{equ:minimum}
T_{\text{opt}} = \left( \frac{L}{\left( \frac{\beta C_T}{\alpha C_M} \right)^{\frac{1}{1-\alpha}}} \right)^{\frac{1-\alpha}{2-\beta-\alpha}}, \text{~~~~~~~}
M_{\text{opt}} = \left( \frac{\beta C_T}{\alpha C_M} \right)^{\frac{1}{1-\alpha}} T^{\frac{1-\beta}{1-\alpha}}
\end{equation}
To verify the effectiveness of this principle, we set $L$ as 6K and obtain $T_{\text{opt}}\approx118$ and $M_{\text{opt}}\approx51$ according to Equation~\ref{equ:minimum}. For the experiment, we vary the number of visual embeddings and frames simultaneously under a fixed visual context length, yielding five $\langle T, M\rangle$ configurations: $\langle 8, 729\rangle, \langle 30, 196\rangle, \langle 72, 81\rangle, \langle 120, 49\rangle, \langle 162, 36\rangle$. For each configuration, we train a video MLLM and calculate the language modeling loss. We also evaluate these models on the long video understanding benchmarks. The results in Table~\ref{tab:fix_window_result} show that the minimum loss is achieved with 120 frames and 49 visual embeddings per frame, which is quite near to $T_{\text{opt}}\approx118$ and $M_{\text{opt}}\approx51$. As for the benchmark evaluation, scaling the number of frames consistently improves overall performance without saturation, even with 162 frames and 36 visual embeddings. This phenomenon occurs because there remains a gap between the next-token-prediction loss and final performance on downstream tasks, but the theoretical minimum point can serve as a strong starting point for subsequent optimization.

\setlength{\tabcolsep}{3.3pt} 
\begin{table}[h]
\caption{Model performance under the same number of total visual embeddings.}
    \small
    \centering
    \begin{tabular}{ccc|c|cccccc}
    \toprule
\makecell[l]{\# Frames} & \makecell[c]{\# Embed./\\Frames}&    \makecell[l]{\# Total\\Embed.}&Loss $\downarrow$&\makecell[c]{Event-\\Bench}&  \makecell[c]{VNBench}& \makecell[c]{MLVU}&  \makecell[c]{LongVideo\\Bench} &\makecell[c]{VideoMME\\wo/w-subs}   &Avg.\\
    \midrule
        8 &729&    5832 &0.642&18.33&  16.30&  50.98&  43.44& 46.22/53.67 &38.16\\
        30&196&   5880 &0.648&28.67& 31.11& 54.97&  48.90&53.19/60.19 &46.17\\
        72&81&   5832 &0.648&24.33& 37.56& 58.37&  \textbf{50.34}&53.04/61.11 &47.46\\
        120&49&    5880 &\textbf{0.639}&\underline{29.67}&  \underline{38.44}&  \underline{59.06}&  49.81&\underline{55.15}/\underline{61.67} &\underline{48.97}\\
        162&36&    5832 &0.653&\textbf{33.00}&  \textbf{40.67}&  \textbf{62.83}&  \underline{50.04}&\textbf{55.19}/\textbf{62.00} &\textbf{50.62}\\
     \bottomrule
    \end{tabular}
    \label{tab:fix_window_result}
\end{table}

\begin{table}[t]
\caption{Experiment results on representative long video understanding benchmarks. \ignore{*LLaVA-OV-72B shorts for LLaVA-OneVision-72B.}}
    \small
    \centering
    \begin{tabular}{lcccccc}
    \toprule
\makecell[l]{Models}&   \makecell[c]{Training\\Data}&\makecell[c]{Event-\\Bench}&  \makecell[c]{VNBench}&  \makecell[c]{VideoMME\\wo/w-subs}& \makecell[c]{MLVU}&  \makecell[c]{LongVideoBench}\\
    \midrule
    \multicolumn{7}{c}{\textit{Proprietary MLLMs}}\\
    \midrule
        GPT-4o&  Unknown&\underline{37.33}&  \underline{64.4}&  \underline{71.9}/\underline{77.2}&  \textbf{64.6}& \textbf{66.7}\\
        Gemini-1.5-Pro& Unknown&\textbf{38.67}& \textbf{66.7}& \textbf{75.0}/\textbf{81.3}& -& \underline{64.0}\\
        GPT-4V& Unknown&27.00 & 48.9& 59.9/63.3& \underline{49.2}& 59.1\\
        Qwen-VL-Max&  Unknown&-& -& 51.3/51.2& 42.2& -\\
    \midrule
    \multicolumn{7}{c}{\textit{Open-source MLLMs}}\\
    \midrule
        Video-CCAM-14B&   4.4M&-&  -&  \underline{53.2}/\underline{57.4}&  \underline{63.1}& -\\
        Video-LLaVA-7B&  2M&5.87& 12.4& 39.9/41.6& 47.3& -\\
        LLaMA-VID-long-7B&   1.6M&0.00&  10.8&  -&  33.2& -\\
        MovieChat-7B&  Unknown&\underline{20.33}& -& -& 25.8& -\\
        VideoChat2-7B&  29M&14.67& 12.4& 39.5/43.8& -& -\\
        ST-LLM-7B&   Unknown&16.67&  \underline{22.7}&  37.9/42.3&  -& -\\
        VideoLLaMA2-7B&   13.4M&-&  -&  46.6/-&  48.5& -\\
        LongVA-7B&   1.3M&-&  -&  52.6/-&  56.3& -\\
        LongViLA-8B&  Unknown&-& -& 50.5/-& -& -\\
        Opt-Visor-7B (Ours)&   2.6M&\textbf{32.00}&  \textbf{45.41}&  \textbf{54.9}/\textbf{62.0}&  \textbf{63.6}& \textbf{51.63}\\
    
    \bottomrule
    \end{tabular}
    \label{tab:main_exp}
\end{table}

\paragraph{Comparison with Existing MLLMs.} We compare our model with a series of representative MLLMs, including four proprietary MLLMs: GPT-4o~\citep{gpt-4o}, Gemini-1.5-Pro~\citep{Reid-Gemini1.5-2024}, GPT-4V~\citep{Openai-GPT-4V-2023}, Qwen-VL-Max~\citep{bai-2023-arxiv-qwenvl}, as well as nine open-source MLLMs: Video-CCAM~\citep{fei2024video}, Video-LLaVA~\citep{lin2023video}, LLaMA-VID-long~\citep{li2023llamavid}, MovieChat~\citep{song2024moviechat}, VideoChat2~\citep{li2024mvbench}, ST-LLM~\citep{liu2025st}, VideoLLaMA2~\citep{cheng2024videollama}, LongVA~\citep{zhang2024longva}, and LongViLA~\citep{xue2024longvila}. To improve the performance of our model, we apply several techniques during training. Specifically, we represent each image in the image instruction with the AnyRes encoding scheme~\citep{zhang2024llavanextvideo}, where each image is divided into multiple tiles to simulate a short video. Additionally, to equip the model with the temporal understanding ability, we introduce a temporal prompt before each frame, resulting in the video representation \emph{``1 <frame 1> 2 <frame 2> ... n <frame n>''}, where \emph{<frame i>} represent the visual embeddings of the $i$-th frame.  Besides, we increase the global batch size from 64 to 128 and train the model for 2 epochs. Under this configuration, we re-train the model with $\langle T, M\rangle = \langle 120, 49 \rangle$ and name it \textbf{Opt-Visor}. We evaluate Opt-Visor and the baseline models on the long video understanding benchmarks. The results in Table~\ref{tab:main_exp} show that Opt-Visor achieves the best performance among all the open-source models, with only 2.6M training samples. Notably, our model even outperforms GPT-4V on certain benchmarks, such as Event-Bench and MLVU, demonstrating the effectiveness of our optimal visual context representation scheme.
\section{Related Work}

\paragraph{Scaling Law.} In the field of LLMs, many existing works have demonstrated that scaling the size of model parameters and training data can consistently improve the model capacity~\citep{radford2019language,brown2020language,touvron2023llama}. As a result, it is necessary to build a quantitative relationship between these scaling factors and the final performance, which is called the scaling law. Two representative scaling laws for LLM are proposed by \citet{kaplan2020scaling} and \citet{hoffmann2022training}, where the former one models the relationship between the loss and model size, dataset size, and the amount of computation budget independently, and the follower one models the relationship between loss and model size, dataset size simultaneously. Inspired by these works, several studies show that the scaling law also holds for different model architectures~\citep{clark2022unified}, training strategies~\citep{gao2023scaling}, and can be transferred to other domains like information retrieval~\citep{fang2024scaling,ardalani2022understanding}, computer vision~\citep{zhai2022scaling,dehghani2023scaling}, and multi-modal~\citep{radford2021learning,alayrac2022flamingo}. 

\paragraph{Video MLLM.} Training an MLLM with long-context video understanding ability is a challenging task and remains underexplored. One line of work focuses on enabling long video training from the perspectives of training systems~\citep{xue2024longvila}, training strategies~\citep{liu2024kangaroo}, and model architectures~\citep{wang2024longllava}. For example, LongVILA~\citep{xue2024longvila} proposes the first Multi-Modal Sequence Parallelism system for long-context training and inference. Kangaroo~\citep{liu2024kangaroo} utilizes a curriculum training pipeline to gradually increase the number of frames during training. LongLLaVA~\citep{wang2024longllava} adapts the model architecture to a hybrid of Mamba~\citep{gu2023mamba} and Transformer~\citep{vaswani2017attention} blocks. Another line of work aims to enable long video understanding during inference~\citep{song2024moviechat, zhang2024longva}. For example, MovieChat~\citep{song2024moviechat} proposes a memory mechanism that includes a rapidly updated short-term memory and a compact long-term memory to store representations of long videos. LongVA~\citep{zhang2024longva} extends the context window of an LLM and demonstrates that long video understanding can be directly transferred from an MLLM without any video-specific training. 

\paragraph{Visual Embedding Compression.} Most existing MLLMs are typically composed of a vision encoder, an LLM, and a visual projector to project the image embeddings into the semantic space of the LLM. Early works like Flamingo~\citep{alayrac2022flamingo} adopt a resampler, which inserts a cross-attention module into the LLM layer to extract visual features, and this is followed by IDEFICS~\citep{laurenccon2024obelics} and Otter~\citep{li2023otter}. Similar to the resampler, BLIP-2~\citep{li2023blip} and InstructBLIP~\citep{dai2023instructblip} utilize a cross-attention module called Q-Former to compress the image embeddings and directly input the resulting visual embeddings into the LLM. Another line of work, represented by LLaVA~\citep{liu2024visual}, directly projects the image embeddings into the semantic space of the LLM with an MLP, achieving decent performance and converging quickly. Based on this, several works~\citep{yao2024deco, cai2024matryoshka} propose adding a pooling module after the MLP to reduce the number of visual embeddings. However, when adapting MLLMs to video tasks, the design of the projector is less explored. LLaVA-NeXT-Video~\citep{zhang2024llavanextvideo} and LLaVA-OneVision~\citep{li2024llava} use mean pooling or bilinear interpolation to aggregate visual embeddings, while neglecting the temporal dependency of video frames. To model the temporal dependency, VideoLLaMA2~\citep{cheng2024videollama} introduces a downsampling module and a spatial-temporal convolution module. 

\section{Conclusion}
\label{sec:conclusion}
In this work, we explored the basic design space of visual context representation in video MLLMs from two major aspects, \ie determining the number of sampled frames per video (\emph{frame selection}) and visual embeddings per frame (\emph{embedding selection}). We conducted experiments based on widely-used Video-MLLM architecture using different sampling and compression strategies, and varied the number of visual embeddings and frames to collect the data points.
We first formulated the studied task as a constrained optimization problem, and then studied the scaling effects for frame selection and embedding selection. 
Then, we fitted the performance function curve \emph{w.r.t.} the two factors, and derived several important empirical findings to determine the two factors. 
Finally, we modeled the joint effects of the two factors,  derived the optimal setting, and then verified the effectiveness with empirical experiments. 
Our work has revealed that visual context representation has an important effect on the model performance of video MLLMs, which is worth more research attention. In future work, we will investigate more strategies for frame and embedding selection, and also consider designing new architectures that can support video context representation.  


\section*{Limitation}
In this work, we explore the basic design space of visual context modeling by varying the numbers of sampled frames per video and visual embeddings per frame. Although we also consider using different sampling and compressing strategies, there are also other important designs that should be explored in the future, \eg different backbone LLMs and visual encoders.
Furthermore, we only consider two metrics to study the scaling effect, \ie language modeling loss and zero-shot accuracy on benchmarks. 
Besides, we conduct all the experiments on the classic LLaVA architecture. It is also necessary to test the effectiveness of our conclusion on other architectures.

\bibliography{baichuan}
\bibliographystyle{baichuan}

\newpage
\appendix

\section{Appendix}
The details of the training data are listed in Table~\ref{tab:dataset}.

The training hyperparameters are listed in Table~\ref{tab:hyperparameter}.

The estimated optimal visual tokens and frames for various visual context lengths are listed in table~\ref{tab:configuration}.
\begin{table}[h]
\small
    \centering
    \caption{The statistics of our training data, including 1.8M image-text instructions and 0.7M video-text instructions.}
    \begin{tabular}{l|l|c}
    \toprule
         Modality&  Dataset&   Samples\\
    \midrule
         Image-Text&  Cauldron&  1.8M\\
    \midrule
         \multirow{9}{*}{Video-Text}&  VideoChatGPT-100K&  100K\\
         &  ShareGPT4Video&  40K\\
         &  ShareGPTVideo&  255K\\
         &  VIM&  32K\\
         & NExT-QA&40K\\
         & SthSthV2&40K\\
         & STAR&40K\\
         & TextVR&40K\\
         & CLEVRER&80K\\
        & Kinetics-710&40K\\
    \midrule
    Total&-&2.5M\\
    \bottomrule
    \end{tabular}
    \label{tab:dataset}
\end{table}

\begin{table}[h]
    \centering
    \caption{Training hyperparameters.}
    \begin{tabular}{c|c}
    \toprule
         Hyperparameter& Value\\
    \midrule
         Global batch size& 64\\
         Gradient clipping& 1\\
         Weight decay& 0\\
         Warmup ratio& 0.03\\
         LLM lr&2e-5\\
         Projector lr&1e-4\\
         Vision encoder lr&2e-6\\
         lr schedule&cosine\\
    \bottomrule
    \end{tabular}
    \label{tab:hyperparameter}
\end{table}

\begin{table}[h]
    \centering
    \caption{Estimated Optimal Visual Tokens and Frames for Various Visual Context Lengths.}
    \begin{tabular}{ccc}
         \toprule
         Visual Context Length&  \# Visual Tokens& \# Frames\\
         \midrule
         6,000&  51& 118\\
         14,000&  78& 178\\
         30,000&  116& 258\\
         62,000&  169& 367\\
        126,000& 243&517\\
        \bottomrule
    \end{tabular}
    \label{tab:configuration}
\end{table}
\end{document}